\pdfoutput=1

\documentclass[11pt]{article}

\usepackage{ACL2023} 

\usepackage{times}
\usepackage{latexsym}

\usepackage[T1]{fontenc}

\usepackage[utf8]{inputenc}

\usepackage{microtype}

\usepackage{inconsolata}

\usepackage{caption}

\usepackage{graphicx}

\usepackage{arydshln}
\usepackage{multirow}

\usepackage{booktabs}

\usepackage{amssymb}
\usepackage{pifont}
\newcommand{\cmark}{\ding{51}}%
\newcommand{\xmark}{\ding{55}}%

\usepackage{amsmath}
\usepackage{float}

\title{CL-UZH at SemEval-2023 Task 10: Sexism Detection through \\ Incremental Fine-Tuning and Multi-Task Learning with Label Descriptions}

\author{Janis Goldzycher \\
  Department of Computational Linguistics \\
  University of Zurich  \\
  \texttt{goldzycher@cl.uzh.ch} \\}

\begin{document}
\maketitle
\begin{abstract}
The widespread popularity of social media has led to an increase in hateful, abusive, and sexist language, motivating methods for the automatic detection of such phenomena.
The goal of the SemEval shared task \textit{Towards Explainable Detection of Online Sexism} (EDOS 2023) is to detect sexism in English social media posts (subtask A), and to categorize such posts into four coarse-grained sexism categories (subtask B), and eleven fine-grained subcategories (subtask C). 
In this paper, we present our submitted systems for all three subtasks, based on a multi-task model that has been fine-tuned on a range of related tasks and datasets before being fine-tuned on the specific EDOS subtasks.
We implement multi-task learning by formulating each task as binary pairwise text classification, where the dataset and label descriptions are given along with the input text. 
The results show clear improvements over a fine-tuned DeBERTa-V3 serving as a baseline leading to $F_1$-scores of 85.9\% in subtask A (rank 13/84), 64.8\% in subtask B (rank 19/69), and 44.9\% in subtask C (26/63).\footnote{We make our code publicly available at \url{https://github.com/jagol/CL-UZH-EDOS-2023}.}

{\color{red}OFFENSIVE CONTENT WARNING: This report contains some examples of hateful content. This is strictly for the purposes of enabling this research, and we have sought to minimize the number of examples where possible. Please be aware that this content could be offensive and cause you distress.}
\end{abstract}

\section{Introduction}
With social media's expanding influence, there has been a rising emphasis on addressing the widespread issue of harmful language, especially sexist language \cite{1633535, simons2015addressing, 10.1145/3427478.3427482}.
Automatic content moderation and monitoring methods have become indispensable due to the sheer amount of posts and comments on social media platforms. 
However, the deployment of automatic methods has led to a new problem: current approaches to sexism detection rely on transformer-based language models whose inner workings, in spite of model interpretability research, generally remain opaque \cite{sun_interpreting_2021}. This stands in contrast with the need for explainable and transparent decision processes in content moderation.

The EDOS 2023 shared task \cite{kirkSemEval2023} focuses on the detection (subtask A), and coarse- (subtask B) and fine-grained (subtask C) categorization of sexism. 
The purpose of sexism categorization is to aid the explainability of sexism detection models, where categorization can serve as additional information for why a post was classified as sexist.

In this paper, we present our approach for all three subtasks. 
The annotated data for detecting sexism is scarce compared to other natural language processing tasks and is often not publicly available. In response to this, we adopt a multi-task learning approach, where we first train a general model for the detection of hateful and abusive language, and incrementally adapt it to the target task.

\begin{figure*}[ht]
\center
\includegraphics[width=0.9\linewidth]{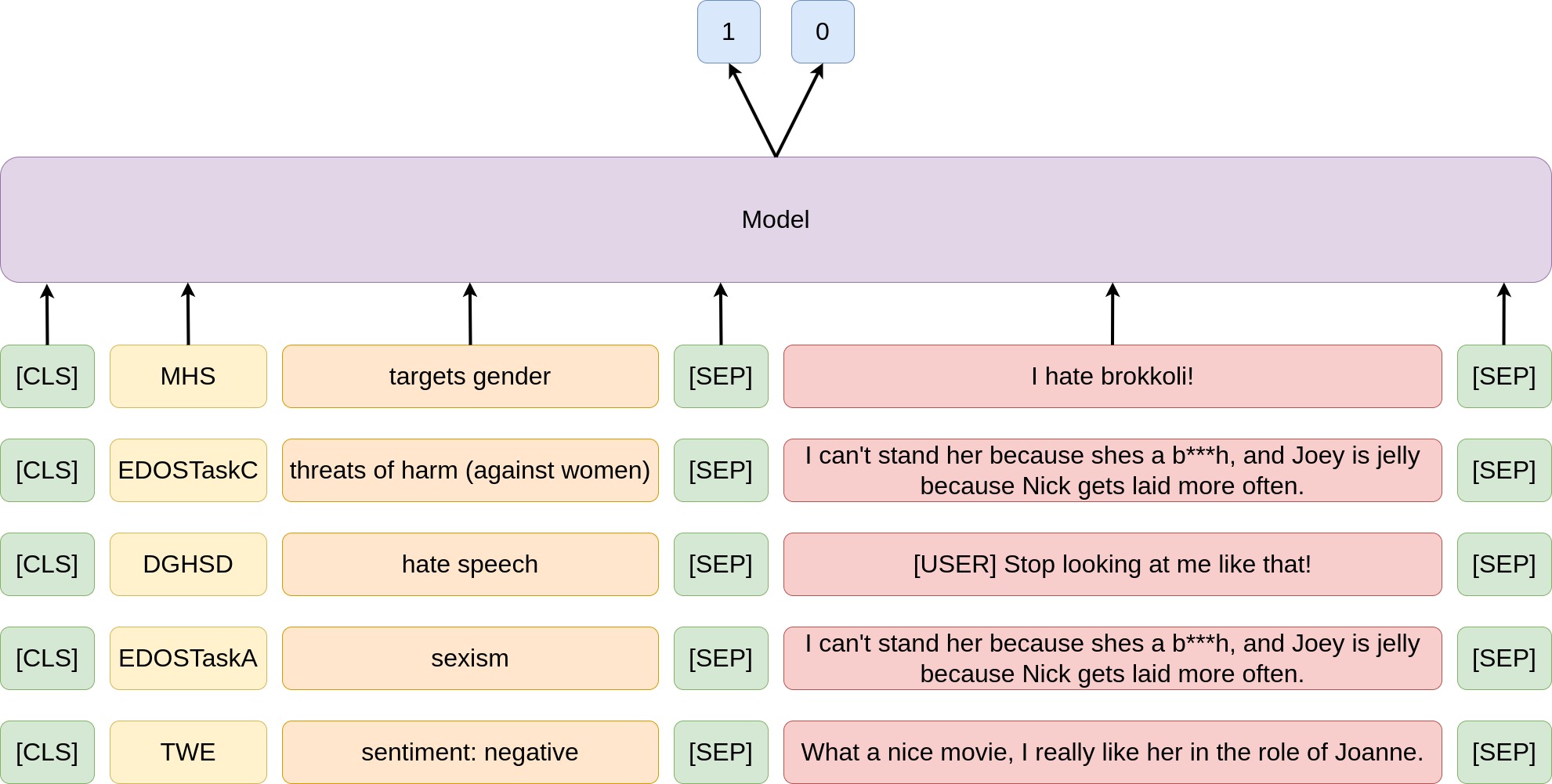}
\caption{Task Formulation: The task is formulated as binary pairwise text classification where the model receives as input a dataset identifier, a label description, and an input text and predicts if the label, as learned for the given dataset, applies to the input text. Note that the same input text can appear with different annotations.}
\label{fig:label-descriptions}
\end{figure*}

We implement multi-task learning via manipulation of the input, concretely by adding label descriptions, and dataset identifiers. This means that the model is presented with a pairwise text classification task where it gets a label description and a dataset identifier as a first sequence and the text to classify as the second sequence.
The model then learns to predict if the label description presented in the first sequence, in the context of a dataset identifier, applies to the input text of the second sequence.
Figure \ref{fig:label-descriptions} demonstrates the approach. 
We collect data for a range of related tasks, including hate speech detection, offensive language detection, emotion detection, stance detection on feminism and abortion, and target group detection, leading to an auxiliary training set of over 560,000 annotated examples.

Our method involves a three-stage training process. As a first step, we train a general abusive language detection model using all available training data. In the second step, we further fine-tune this model on all three EDOS subtasks, and finally, in the third step, we fine-tune the model only on the target subtask. 

Our models obtain strong results for subtask A, a macro-$F_1$ score of 0.859 achieving place 13 out of 84, but rank lower in subtasks B and C, indicating the proposed approach works comparatively well with few classes during inference time, but decreases in performance, relative to other approaches, with a higher number of classes. 
Our ablation study demonstrates that multi-task learning with label descriptions leads to clear performance improvements over a baseline consisting of DeBERTa-V3 \cite{he2021debertav3} fine-tuned on each subtask. However, it remains unclear if there is a positive contribution from the additionally proposed dataset identifier.

\begin{figure*}[t]
\center
\includegraphics[width=\linewidth]{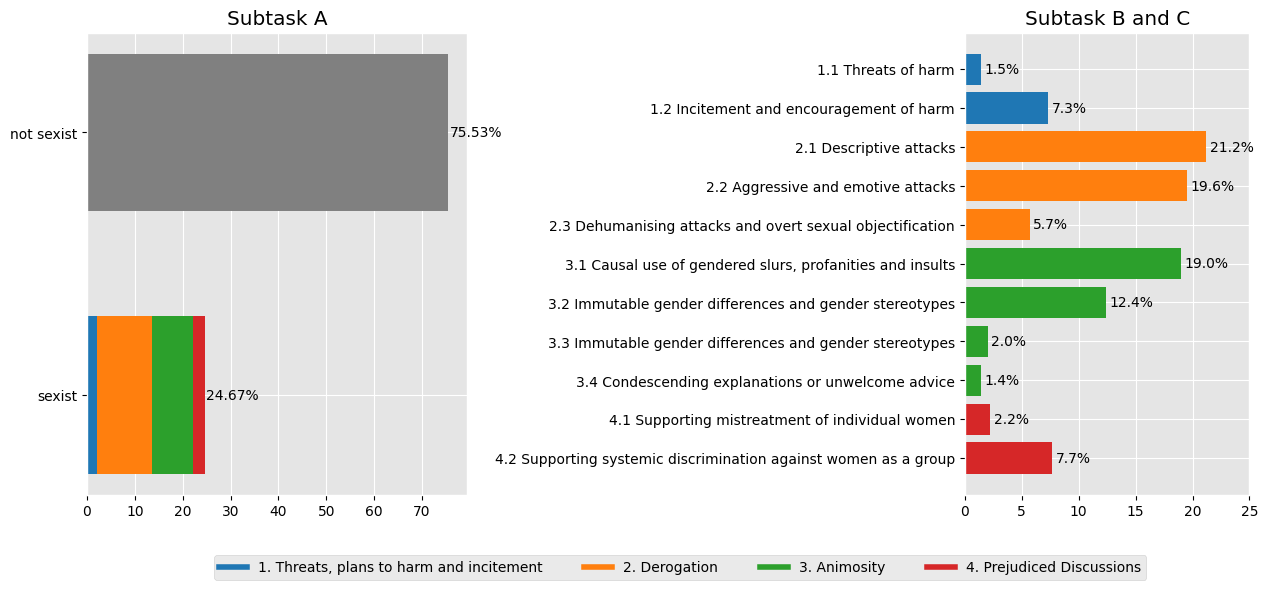}
\caption{EDOS class distribution. Note that the category \textit{not sexist} is absent from subtasks B and C. The percentages provided for these subtasks pertain solely to the \textit{sexist} class, rather than the entire dataset.}
\label{fig:EDOS-class-distribution}
\end{figure*}

\section{Related Work}
\subsection{Sexism Detection}
Sexism detection, sometimes also called sexism identification, is the task of predicting if a given text (typically a social media post) is sexist or not.
Most research on the detection of harmful language has focused on more general phenomena such as offensive language \cite{pradhan_review_2020}, abusive language \cite{nakov_detecting_2021}, or hate speech \cite{10.1145/3232676}. Sexism intersects with these concepts but is not entirely covered by them since it also refers to subtle prejudices and stereotypes expressed against women.
Accordingly, datasets for hate speech often include women as one of multiple target groups \cite{mollasETHOSMultiLabelHate2022, vidgen-etal-2021-learning, waseem-2016-racist}, and thus contain sexist texts, but are not exhaustive of sexism, since they do not cover its subtle forms.
Recently, there has been increased attention on the detection of sexism and misogyny, leading to one shared task on sexism detection \cite{RodrguezSnchez2021OverviewOE} and three shared tasks on misogyny detection \cite{fersini2018ami, fersini2018ibereval, fersini2020ami}. 

Harmful language detection tasks, such as sexism detection, are typically formulated as binary text classification tasks \cite{10.1145/3232676}. 
Categorizing sexism and misogyny is usually cast as a single-label multi-class classification task \cite{fersini2018ami, fersini2018ibereval} with the exception of \citet{parikh-etal-2019-multi} who formulate the task as multi-label multi-task classification. 
Earlier approaches to sexism detection varied in their methods ranging from manual feature engineering using \textit{n}-grams, lexical features, and word embeddings with statistical machine learning \cite{fersini2018ami, fersini2018ibereval} to custom neural architectures \cite{fersini2020ami}. Current approaches typically rely on fine-tuning pre-trained language models \cite{fersini2020ami, RodrguezSnchez2021OverviewOE}. 

\subsection{Label descriptions and Multi-Task Learning}
Prompts \cite{liu_pre-train_2021}, and task descriptions \cite{raffel_exploring_2020} have been used to condition generative models, while hypotheses \cite{wang_entailment_2021} and label descriptions \cite{Zhang2022LongtailedEM} have been used to condition classification models in multi-task settings to produce a desired output.
Multiple works have shown that multi-task learning \cite{Caruana1998, ruder2017overview} with auxiliary tasks such as polarity classification, aggression detection, emotion detection, and offensive language detection can benefit sexism detection \cite{abburi-etal-2020-semi,plaza2021sexism,rodriguez2021multi,safi-samghabadi-etal-2020-aggression}.
However, to the best of our knowledge, our approach is the first to implement multi-task learning for sexism detection and categorization via label descriptions and without multiple model heads.

\section{Data}

\subsection{EDOS Dataset}
The EDOS 2023 dataset \cite{kirkSemEval2023} contains 20,000 posts from Gab and Reddit, labelled on three levels. On the top level, it is annotated with the binary labels \textit{sexism} and \textit{not-sexism}, where \textit{sexism} is defined as ``any abuse or negative sentiment that is directed towards women based on their gender or based on their gender combined with one or more other identity attributes (e.g. Black women, Muslim women, Trans women).''.\footnote{\url{https://codalab.lisn.upsaclay.fr/competitions/7124\#learn_the_details-overview}
}
The posts labelled as sexist are further classified into one of four categories, and eleven subcategories called vectors. The label taxonomy and the respective class distributions are displayed in Figure \ref{fig:EDOS-class-distribution}.

14,000 labelled examples are released as training data, and 2,000 examples and 4,000 examples are held back for validation and testing, respectively. 
Additionally, the shared task organizers provide one million unlabelled examples from Reddit and one million unlabelled examples from Gab. For our approach, we do not make use of the unlabelled data. 

\begin{table}[t]
\centering
\small
\begin{tabular}{lllr}
\toprule
dataset & label type & label value & \multicolumn{1}{l}{size} \\
\midrule
DGHSD & hate speech & \begin{tabular}[c]{@{}l@{}}yes: 46.1\% \\ no: 53.9\%\end{tabular} & 32,924 \\
\hdashline
SBF & lewd & \begin{tabular}[c]{@{}l@{}}yes: 10.1\%\\ no: 89.9\%\end{tabular} & 35,424 \\
 & offensive & \begin{tabular}[c]{@{}l@{}}yes: 47.1\%\\ no: 52.9\%\end{tabular} & 35,424 \\
 \hdashline
MHS & hate speech & \begin{tabular}[c]{@{}l@{}}yes: 40.5\%\\ no: 59.5\%\end{tabular} & 130,000 \\
 & targets gender & \begin{tabular}[c]{@{}l@{}}yes: 29.8\%\\ no: 70.2\%\end{tabular} & 130,000 \\
 & targets women & \begin{tabular}[c]{@{}l@{}}yes: 21.9\%\\ no: 78.1\%\end{tabular} & 130,000 \\
 \hdashline
TWE & offensive & \begin{tabular}[c]{@{}l@{}}yes: 33.1\%\\ no: 66.9\%\end{tabular} & 11,916 \\
 & sentiment & \begin{tabular}[c]{@{}l@{}}negative: 15.5\%\\ neutral: 45.3\%\\ positive: 39.1\%\end{tabular} & 45,615 \\
 & emotion & \begin{tabular}[c]{@{}l@{}}anger: 43.0\% \\ joy: 21.7\%\\ optimism: 9.0\%\\ sadness: 26.3\%\end{tabular} & 3,257 \\
 & hate & \begin{tabular}[c]{@{}l@{}}yes: 42.0\%\\ no: 58.0\%\end{tabular} & 9,000 \\
 & irony & \begin{tabular}[c]{@{}l@{}}yes: 50.5\%\\ no: 49.5\%\end{tabular} & 2,862 \\
 & stance feminist & \begin{tabular}[c]{@{}l@{}}none: 18.9\%\\ against: 49.4\%\\ favor: 31.7\%\end{tabular} & 597 \\
 & stance abortion & \begin{tabular}[c]{@{}l@{}}none: 27.1\%\\ against: 54.3\%\\ favor: 18.6\%\end{tabular} & 587 \\
 \bottomrule
\end{tabular}
\caption{\label{tab:aux-datasets}
Label distributions of the auxiliary datasets.
}
\vspace{-0.5cm}
\end{table}

\subsection{Auxiliary Datasets}

However, we do make use of the following additional, labelled datasets as auxiliary training sets for multi-task learning:

\paragraph{DGHSD} The ``Dynamically Generated Hate Speech Dataset'' \cite{vidgen-etal-2021-learning} contains artificial adversarial examples aimed at tricking a binary hate speech detection model into predicting the wrong class.

\paragraph{MHS} The ``Measuring Hate Speech'' dataset \cite{kennedy_constructing_2020} contains comments sourced from Youtube, Twitter, and Reddit and is annotated for ten attributes related to hate speech. We only use the subset of labels listed in Table \ref{tab:aux-datasets}. 

\paragraph{SBF} The ``Social Bias Frames'' dataset \cite{sap-etal-2020-social} is a combination of multiple Twitter datasets \cite{founta_large_2018, Davidson_Warmsley_Macy_Weber_2017, waseem_hateful_2016} with newly collected data from Reddit, Gab, and Stormfront. We only use the subset of labels listed in Table \ref{tab:aux-datasets}. 

\paragraph{TWE} ``TweetEval'' \cite{barbieri-etal-2020-tweeteval} combines multiple datasets for different tasks into a single benchmark for detecting various aspects of tweets. We use the datasets for emotion classification \cite{mohammad2018semeval}, irony detection \cite{van-hee-etal-2018-semeval}, hate speech detection \cite{basile-etal-2019-semeval}, offensive language detection \cite{zampieri2019semeval}, sentiment detection \cite{rosenthal-etal-2017-semeval}, and stance detection \cite{mohammad2016semeval} for stances on the topics \textit{feminism} and \textit{abortion}.

\subsection{Preprocessing}
During preprocessing, we replaced all URLs in the input texts with the placeholder string ``\texttt{[URL]}'', all usernames (strings starting with an ``\texttt{@}'') with ``\texttt{[USER]}'', and all emojis with the respective textual description, also surrounded by brackets.

\section{System Description}

We formulate each EDOS subtask as a binary pairwise classification task where the model predicts if a given label applies to the input text.
This allows us to simultaneously train on multiple datasets with different labeling schemes and a different number of distinct labels without adjusting the model architecture or having to use multiple model heads. 

Formally, our model receives as input (1) the concatenation of a dataset identifier $d_i \in D$ and a label description $l_j \in L$, and (2) the input text $t$. It predicts the probability distribution $y = \text{softmax}(\text{model}(\text{concat}(d_i, l_j),t))$ where $y \in \mathbb{R}^2$. $y_1$ then denotes the probability that $l_j$, given the context of $d_i$, does apply to $t$. 

\begin{table}[t]
\resizebox{0.5\textwidth}{!}{
\begin{tabular}{llr}
\toprule
\multicolumn{1}{c}{phase} & \multicolumn{1}{c}{Parameter} & \multicolumn{1}{c}{Value} \\
\midrule
\multirow{9}*{general} & loss function & cross-entropy loss \\
 & optimizer & Adam \cite{DBLP:journals/corr/KingmaB14} \\
 & $\beta_1$  & 0.9 \\
 & $\beta_2$ & 0.999 \\
\multicolumn{1}{l}{} & learning rate & 1e-6 \\
\multicolumn{1}{l}{} & warmup steps & 1,000 \\
\multicolumn{1}{l}{} & effective batch size & 32 \\
 & evaluation metric & macro-$F_1$ \\
 & early stopping & \cmark \\
 \hdashline
1/3: AUX + EDOS & max epochs & 1 \\
\hdashline
\multirow{2}*{2/3: EDOS} & max epochs & 20 \\
 & patience & 5 \\
\hdashline
\multirow{2}*{3/3: EDOS A/B/C} & max epochs & 20 \\
 & patience & 5 \\
\bottomrule
\end{tabular}
}
\caption{\label{tab:hyper-parameters}Training hyperparameters. The left column refers to the different training phases. \textit{general} applies to all training phases and all EDOS subtasks. \textit{AUX+EDOS} refers to training on all auxiliary datasets, \textit{EDOS} to training on all EDOS subtasks and \textit{EDOS A/B/C} to training only on the target subtask.}
\end{table}

\subsection{Model Details}
\label{subsec:model-details}
We use DeBERTa \cite{he2020deberta}, specifically DeBERTa-V3-large \cite{he2021debertav3} fine-tuned on a range of natural language inference (NLI) datasets \cite{laurer2022less}\footnote{The model is publicly available at \url{https://huggingface.co/MoritzLaurer/DeBERTa-v3-large-mnli-fever-anli-ling-wanli}.}, since this model is already fine-tuned to classify and relate text pairs. We only change the output dimensionality from 3 to 2 for binary classification. In ablation tests, we also use the DeBERTa-V3-large without further fine-tuning.\footnote{The model is publicly available at \url{https://huggingface.co/microsoft/deberta-v3-large}.}

\subsection{Label Descriptions}

Where possible, we use the label names listed in Figure \ref{fig:EDOS-class-distribution} and Table \ref{tab:aux-datasets} as the label description. However, we make the following exceptions and adjustments:
We strip the numbering from the label names for EDOS subtasks B, and C, and add the string `` (against women)'' at the end since this target group information is not yet in the label name. 
For multi-class classification in auxiliary datasets, we follow the format ``\texttt{<label type>: <label value>}''. For example, for sentiment classification, which has the three possible label values \texttt{negative}, \texttt{neutral}, and \texttt{positive}, we can generate a true example for positive sentiment with the label description ``\texttt{sentiment: positive}''.

\begin{table*}[t]
\centering
\begin{tabular}{lrrrrrrrr}
\toprule
& \multicolumn{1}{c}{A} & \multicolumn{1}{c}{B} & \multicolumn{1}{c}{C} & & \multicolumn{1}{c}{AVG} \\
\midrule
single task EDOS & 0.840 & 0.202 & 0.122 & & 0.388 \\
+ label description & 0.851 & 0.160 & 0.098 & & 0.370 \\
 & \multicolumn{1}{l}{} & \multicolumn{1}{l}{} & \multicolumn{1}{l}{} & & \multicolumn{1}{l}{} \\
multi-task EDOS via label descriptions & 0.851 & 0.504 & 0.248 & & 0.534 \\
+ NLI fine-tuning & 0.854 & 0.556 & 0.352 & & 0.587 \\
+ single task fine-tuning & 0.850 & 0.623 & 0.412 & & 0.629 \\
+ fine-tuning on AUX & 0.858 & 0.633 & 0.417 & & 0.636 \\
+ dataset identifier & 0.858 & 0.629 & 0.431 & & 0.640 \\
+ class balancing & \multicolumn{1}{c}{-} & 0.642 & 0.466 & & \multicolumn{1}{c}{-} \\
\bottomrule
\end{tabular}
\caption{Results of the ablation study on the test set. The metric is macro-$F_1$.}
\label{tab:ablation-results}
\end{table*}

\subsection{Dataset Identifier}

The same labels may have slightly different definitions in different datasets or may be differently applied due to different annotators. If no further information is given to the model, this could be a source of noise.
Multiple datasets contain the label \textit{hate speech} for our auxiliary datasets. 
To account for this, we introduce dataset identifiers, which are short dataset abbreviations of a few characters in length that are concatenated with the label description.

\subsection{Training Procedure}

We train the model in three phases: In the first phase, the model is trained with all available examples of all collected datasets. In the second phase, the best checkpoint from the previous phase is further fine-tuned on EDOS data from all three task levels (subtasks A, B, and C). Finally, in the third phase, the model is fine-tuned only on examples from the relevant subtask. 
We consider all three annotations from the three subtasks per example for validation during the first two phases, resulting in 6,000 annotations for each validation. In the last training phase, the model is only fine-tuned on one subtask. Thus, we only validated on the labels for that specific subtask. 
Further training details are provided in Table \ref{tab:hyper-parameters}.

\subsection{Random Negative Sampling}
When converting a multi-class classification task (such as subtask B and C) to a binary pairwise text classification task, each positive example for class $c_k \in C$, where $k \in \left[0,...,|C|\right]$,
can be turned into $|C| - 1$ negative examples by choosing a label $c_k \in C \setminus \{c_k\}$. However, generating all possible negative examples for a positive example would result in an imbalanced training set. Therefore, in settings with more than two classes, we instead sample a random wrong class label during training to create one negative example for each positive example.\footnote{This applies to EDOS subtasks B and C, and the sentiment- emotion- and stance- detection tasks in TweetEval.} This means the model will be trained on different negative examples in each epoch while the positive examples stay the same.

\subsection{Inference} 
During inference, we predict a probability $p_i$ for each candidate class $c_i \in C$ and select the class with the highest probability. This means that we perform $|C|$ number of forward passes per prediction, except for binary classification (subtask A), where we can use just one forward pass to predict a probability for the label \textit{sexism}.

Since our model produces just one probability for subtask A, we can select a probability threshold. For our official submission and for our ablation experiments, we test the thresholds $\{0.5, 0.6, 0.7, 0.8, 0.9\}$ on the validation set and use the highest performing threshold for the test set.

\begin{table}[t]
\centering
\begin{tabular}{lcc}
\toprule
 & \multicolumn{1}{c}{$F_1$} & \multicolumn{1}{l}{Rank} \\
 \midrule
Subtask A & 0.859 & 13/84 \\
Subtask B & 0.648 & 19/69 \\
Subtask C & 0.449 & 26/63 \\
\bottomrule
\end{tabular}
\caption{Results of the official evaluation on the test set.}
\label{tab:official-results}
\end{table}

\section{Experiments and Results}

Table \ref{tab:official-results} contains the official evaluation scores showing strong results for subtask A and moderately good results for subtasks B and C.

\subsection{Ablation Study}
To illustrate the relative importance of the proposed methods, we systematically add components to a baseline model until we arrive at the submitted models and run each model version with three random seeds for our ablation tests. We evaluate the following settings:
(1) "\textit{single task EDOS}": We start with DeBERTa-V3-large models, fine-tuned on each subtask individually, serving as our baseline. 
(2) "\textit{+ label description}": We add label descriptions while still only training each model on one subtask.
(3) "\textit{multi-task EDOS via label descriptions}": The models are trained on all three subtasks simultaneously using label descriptions.
(4) "\textit{+ NLI fine-tuning}": We repeat the setting but start training from the DeBERTa-V3 checkpoint fine-tuned on NLI datasets (see Section \ref{subsec:model-details}). 
(5) "\textit{+ single task fine-tuning}": We add a second fine-tuning phase in which the multi-task model is only fine-tuned on the target subtask. 
(6) "\textit{+ fine-tuning on AUX }": We add fine-tuning on auxiliary tasks and EDOS simultaneously as a first training phase.
(7) "\textit{+ dataset identifier}": We add the dataset identifier to the input.
(8) \textit{"class balancing"}: Finally, we perform upsampling to increase the relative frequency of scarce classes.\footnote{In subtask B, we increase classes with a frequency below 19\% to \textasciitilde19\%. For subtask C, we upsample classes below 9\% to \textasciitilde9\%.} The upsampled version of the dataset is only used during the last fine-tuning phase.

Table \ref{tab:ablation-results} contains the test set results averaged over the three runs with different seeds. 
The full results for each run, including evaluations after intermediary training phases, are displayed in Appendix \ref{sec:app:full-results}. 
In what follows we analyze the effects of different system components and settings.

\paragraph{Baseline}
The baseline \textit{single task EDOS} shows already a strong performance on subtask A, but leads to surprisingly low scores on subtasks B and C. We assume that this is due to underprediction and low performance of the very scarce classes (four classes of subtask C are below 3\%), which can drastically reduce the macro-$F_1$ score.

\paragraph{Multi-Task Learning on all EDOS-Subtasks} 
Comparing the baseline, with \textit{multi-task EDOS via label descriptions}, shows a clear improvement of 1.1 percentage points (pp) from multi-task learning with label descriptions on subtask A, and drastic improvements, more than doubling performance, for subtasks B and C. Looking at single task models with label descriptions (\textit{+ label description}) reveals that on subtask A the entire increase in performance is due to label descriptions while in subtasks B and C the dramatic performance increases are due to multi-task learning.

\paragraph{Starting with an NLI Model}
Starting training from a DeBERTa-V3 checkpoint that is already fine-tuned on NLI increases scores on all three subtasks. The more classes the task has, the larger is the increase. 

\paragraph{Additional Single-Task Fine-Tuning}
Adding a second subtask-specific fine-tuning phase after training on all EDOS subtasks leads to increases of 6.7pp and 4.0pp for subtasks B and C. However, it reduces performance on subtask A by 0.4pp. 

\paragraph{Multi-Task Learning on Auxiliary Tasks}
Inserting a first training phase that includes all auxiliary tasks and EDOS subtasks into the training process leads to further improvements of up to 1.0pp on all subtasks.

\paragraph{The Dataset Identifier}
Adding a dataset identifier to the input leads to mixed results. On subtask A the model does not change in performance, on subtask B it slightly decreases, and on subtask C we observe a clear increase of 1.4pp. Overall, we cannot draw a clear conclusion about the effects of the dataset identifier.

\paragraph{Class Balancing}
Finally, we observe that upsampling low-frequency classes in subtasks B and C has positive effects of 1.3pp and 3.5pp respectively. 

\begin{figure*}[t]
\center
\includegraphics[width=0.6\linewidth]{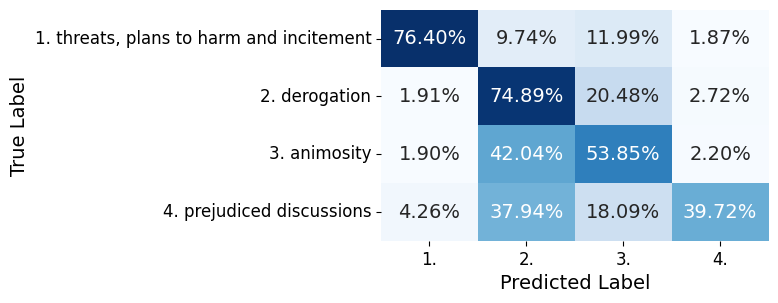}
\caption{Visualized normalized confusion matrix for subtask B.}
\label{fig:taskb-conf-matrix}
\end{figure*}

\begin{figure*}[t]
\center
\includegraphics[width=\linewidth]{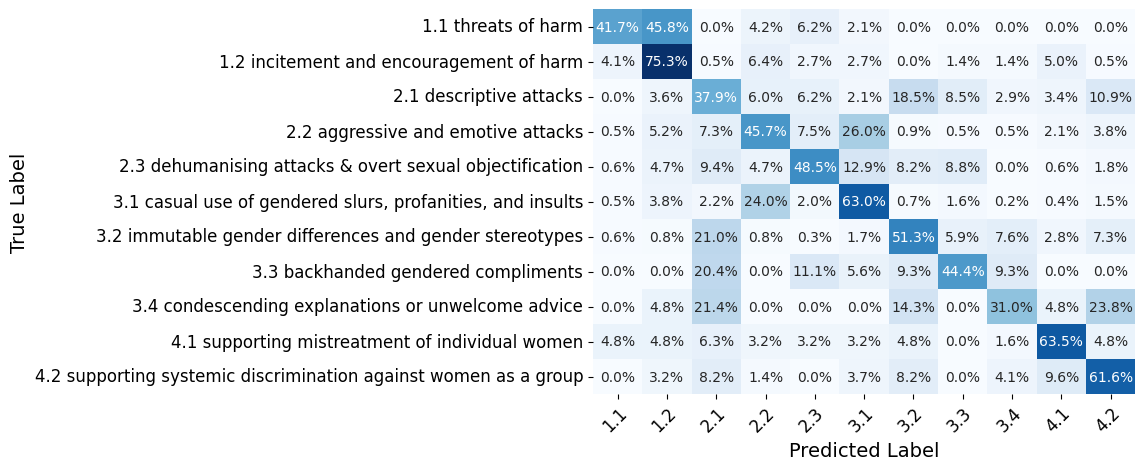}
\caption{Visualized normalized confusion matrix for subtask C.}
\label{fig:taskc-conf-matrix}
\end{figure*}

\subsection{Error Analysis}

Figures \ref{fig:taskb-conf-matrix} and \ref{fig:taskc-conf-matrix} display the confusion matrices averaged over three random seeds for the submitted model configurations for subtasks B and C.

In subtask B, we see that the category \textit{threats, plans to harm, and incitement} is accurately predicted. The \textit{derogation} class has a high recall, likely because it is the most common category. However, this also results in a significant number of false positives from the \textit{animosity} and \textit{prejudiced discussions} classes. As a consequence, these last two classes are strongly underpredicted.

In subtask C, it is evident that mispredictions generally stem from class confusions within the subtask B categories. Erroneous predictions outside of these categories are uncommon, except for \textit{descriptive attacks}, which are frequently mistaken for various forms of \textit{animosity}.

\section{Conclusion}
In this paper, we presented our approaches and results for all three subtasks of the shared task \textit{Towards Explainable Sexism Detection}. 
We developed and evaluated a multi-task learning model that is trained in three phases: (1) training a general multi-task abusive language detection model, (2) fine-tuning the model on all three EDOS subtasks, thus specializing it in sexism detection, and (3) fine-tuning the model only on the target subtask.
We implemented the multi-task capabilities only via input manipulation, i.e., label descriptions and dataset identifiers, without modifying the model architecture or using multiple model heads.

In the official shared task evaluation, our approach led to strong results on subtask A and moderately good results on subtask B and C, indicating that the method decreases more in performance with a higher number of classes than other approaches.
Our ablation tests demonstrate that multi-task learning via label descriptions led to significant performance improvements on subtask A and large performance improvements on subtasks B and C. 
It remains unclear if the dataset identifier has any positive effect.
Overall the results show that our model for binary sexism detection is reliable, but that there is still much room for improvement in sexism categorization. 

\section*{Acknowledgments}
We thank Chantal Amrhein and Simon Clematide, for the valuable conversations, and suggestions, and Jonathan Schaber, and Gerold Schneider for the helpful comments. We also thank the anonymous reviewers for their constructive feedback.

\bibliography{anthology,references,custom}

\begin{thebibliography}{42}
\expandafter\ifx\csname natexlab\endcsname\relax\def\natexlab#1{#1}\fi

\bibitem[{Abburi et~al.(2020)Abburi, Parikh, Chhaya, and
  Varma}]{abburi-etal-2020-semi}
Harika Abburi, Pulkit Parikh, Niyati Chhaya, and Vasudeva Varma. 2020.
\newblock \href {https://doi.org/10.18653/v1/2020.coling-main.511}
  {Semi-supervised multi-task learning for multi-label fine-grained sexism
  classification}.
\newblock In \emph{Proceedings of the 28th International Conference on
  Computational Linguistics}, pages 5810--5820, Barcelona, Spain (Online).
  International Committee on Computational Linguistics.

\bibitem[{Barbieri et~al.(2020)Barbieri, Camacho-Collados, Espinosa~Anke, and
  Neves}]{barbieri-etal-2020-tweeteval}
Francesco Barbieri, Jose Camacho-Collados, Luis Espinosa~Anke, and Leonardo
  Neves. 2020.
\newblock \href {https://doi.org/10.18653/v1/2020.findings-emnlp.148}
  {{T}weet{E}val: Unified benchmark and comparative evaluation for tweet
  classification}.
\newblock In \emph{Findings of the Association for Computational Linguistics:
  EMNLP 2020}, pages 1644--1650, Online. Association for Computational
  Linguistics.

\bibitem[{Basile et~al.(2019)Basile, Bosco, Fersini, Nozza, Patti,
  Rangel~Pardo, Rosso, and Sanguinetti}]{basile-etal-2019-semeval}
Valerio Basile, Cristina Bosco, Elisabetta Fersini, Debora Nozza, Viviana
  Patti, Francisco~Manuel Rangel~Pardo, Paolo Rosso, and Manuela Sanguinetti.
  2019.
\newblock \href {https://doi.org/10.18653/v1/S19-2007} {{S}em{E}val-2019 task
  5: Multilingual detection of hate speech against immigrants and women in
  {T}witter}.
\newblock In \emph{Proceedings of the 13th International Workshop on Semantic
  Evaluation}, pages 54--63, Minneapolis, Minnesota, USA. Association for
  Computational Linguistics.

\bibitem[{Caruana(1998)}]{Caruana1998}
Rich Caruana. 1998.
\newblock \href {https://doi.org/10.1007/978-1-4615-5529-2_5} {\emph{Multitask
  Learning. Learning to Learn}}, pages 95--133. Springer US, Boston, MA.

\bibitem[{Das et~al.(2020)Das, Mathew, Saha, Goyal, and
  Mukherjee}]{10.1145/3427478.3427482}
Mithun Das, Binny Mathew, Punyajoy Saha, Pawan Goyal, and Animesh Mukherjee.
  2020.
\newblock \href {https://doi.org/10.1145/3427478.3427482} {Hate speech in
  online social media}.
\newblock \emph{SIGWEB Newsl.}, Autumn 2020.
\newblock New York, NY, USA.

\bibitem[{Davidson et~al.(2017)Davidson, Warmsley, Macy, and
  Weber}]{Davidson_Warmsley_Macy_Weber_2017}
Thomas Davidson, Dana Warmsley, Michael Macy, and Ingmar Weber. 2017.
\newblock \href {https://doi.org/10.1609/icwsm.v11i1.14955} {Automated hate
  speech detection and the problem of offensive language}.
\newblock \emph{Proceedings of the International AAAI Conference on Web and
  Social Media}, 11(1):512--515.
\newblock Montréal, Québec, Canada.

\bibitem[{Fersini et~al.(2018{\natexlab{a}})Fersini, Nozza, Rosso
  et~al.}]{fersini2018ami}
Elisabetta Fersini, Debora Nozza, Paolo Rosso, et~al. 2018{\natexlab{a}}.
\newblock Overview of the {EVALITA} 2018 {Task} on {Automatic Misogyny
  Identification} (ami).
\newblock In \emph{EVALITA Evaluation of NLP and Speech Tools for Italian
  Proceedings of the Final Workshop 12-13 December 2018, Naples}. Accademia
  University Press.

\bibitem[{Fersini et~al.(2020)Fersini, Nozza, Rosso et~al.}]{fersini2020ami}
Elisabetta Fersini, Debora Nozza, Paolo Rosso, et~al. 2020.
\newblock {AMI@ EVALITA2020: Automatic Misogyny Identification}.
\newblock In \emph{Proceedings of the 7th Evaluation Campaign of Natural
  Language Processing and Speech Tools for Italian (EVALITA 2020)}. CEUR
  Workshop Proceedings. Online.

\bibitem[{Fersini et~al.(2018{\natexlab{b}})Fersini, Rosso, and
  Anzovino}]{fersini2018ibereval}
Elisabetta Fersini, Paolo Rosso, and Maria Anzovino. 2018{\natexlab{b}}.
\newblock {Overview of the Task on Automatic Misogyny Identification at
  IberEval 2018.}
\newblock \emph{Ibereval@ sepln}, 2150:214--228.
\newblock Seville (Spain).

\bibitem[{Fortuna and Nunes(2018)}]{10.1145/3232676}
Paula Fortuna and S\'{e}rgio Nunes. 2018.
\newblock \href {https://doi.org/10.1145/3232676} {{A Survey on Automatic
  Detection of Hate Speech in Text}}.
\newblock \emph{ACM Comput. Surv.}, 51(4).
\newblock New York, NY, USA.

\bibitem[{Founta et~al.(2018)Founta, Djouvas, Chatzakou, Leontiadis, Blackburn,
  Stringhini, Vakali, Sirivianos, and Kourtellis}]{founta_large_2018}
Antigoni~Maria Founta, Constantinos Djouvas, Despoina Chatzakou, Ilias
  Leontiadis, Jeremy Blackburn, Gianluca Stringhini, Athena Vakali, Michael
  Sirivianos, and Nicolas Kourtellis. 2018.
\newblock \href
  {https://www.aaai.org/ocs/index.php/ICWSM/ICWSM18/paper/view/17909} {Large
  {Scale} {Crowdsourcing} and {Characterization} of {Twitter} {Abusive}
  {Behavior}}.
\newblock In \emph{Twelfth {International} {AAAI} {Conference} on {Web} and
  {Social} {Media}}.

\bibitem[{He et~al.(2021)He, Gao, and Chen}]{he2021debertav3}
Pengcheng He, Jianfeng Gao, and Weizhu Chen. 2021.
\newblock Debertav3: Improving deberta using electra-style pre-training with
  gradient-disentangled embedding sharing.
\newblock \emph{arXiv preprint arXiv:2111.09543}.

\bibitem[{He et~al.(2020)He, Liu, Gao, and Chen}]{he2020deberta}
Pengcheng He, Xiaodong Liu, Jianfeng Gao, and Weizhu Chen. 2020.
\newblock Deberta: Decoding-enhanced bert with disentangled attention.
\newblock \emph{arXiv preprint arXiv:2006.03654}.

\bibitem[{Kennedy et~al.(2020)Kennedy, Bacon, Sahn, and von
  Vacano}]{kennedy_constructing_2020}
Chris~J. Kennedy, Geoff Bacon, Alexander Sahn, and Claudia von Vacano. 2020.
\newblock \href {http://arxiv.org/abs/2009.10277} {Constructing interval
  variables via faceted {Rasch} measurement and multitask deep learning: a hate
  speech application}.
\newblock ArXiv:2009.10277 [cs].

\bibitem[{Kingma and Ba(2015)}]{DBLP:journals/corr/KingmaB14}
Diederik~P. Kingma and Jimmy Ba. 2015.
\newblock \href {http://arxiv.org/abs/1412.6980} {Adam: {A} method for
  stochastic optimization}.
\newblock In \emph{3rd International Conference on Learning Representations,
  {ICLR} 2015, San Diego, CA, USA, May 7-9, 2015, Conference Track
  Proceedings}.

\bibitem[{Kirk et~al.(2023)Kirk, Yin, Vidgen, and Röttger}]{kirkSemEval2023}
Hannah~Rose Kirk, Wenjie Yin, Bertie Vidgen, and Paul Röttger. 2023.
\newblock \href {https://doi.org/10.48550/arXiv.2303.04222} {{SemEval}-2023
  {Task} 10: {Explainable} {Detection} of {Online} {Sexism}}.
\newblock In \emph{Proceedings of the 17th International Workshop on Semantic
  Evaluation}, Toronto, Canada. {Association for Computational Linguistics}.

\bibitem[{Laurer et~al.(2022)Laurer, van Atteveldt, Casas, and
  Welbers}]{laurer2022less}
Moritz Laurer, Wouter van Atteveldt, Andreu Casas, and Kasper Welbers. 2022.
\newblock \href {https://osf.io/wqc86/} {{Less Annotating}, {More Classifying}
  – {Addressing} the {Data Scarcity Issue} of {Supervised Machine Learning}
  with {Deep Transfer Learning} and {BERT} - {NLI}}.
\newblock \emph{online}.

\bibitem[{Liu et~al.(2021)Liu, Yuan, Fu, Jiang, Hayashi, and
  Neubig}]{liu_pre-train_2021}
Pengfei Liu, Weizhe Yuan, Jinlan Fu, Zhengbao Jiang, Hiroaki Hayashi, and
  Graham Neubig. 2021.
\newblock \href {http://arxiv.org/abs/2107.13586} {Pre-train, {Prompt}, and
  {Predict}: {A} {Systematic} {Survey} of {Prompting} {Methods} in {Natural}
  {Language} {Processing}}.
\newblock \emph{arXiv:2107.13586 [cs]}.
\newblock ArXiv: 2107.13586.

\bibitem[{Meyer and Cukier(2006)}]{1633535}
R.~Meyer and M.~Cukier. 2006.
\newblock \href {https://doi.org/10.1109/DSN.2006.12} {Assessing the attack
  threat due to irc channels}.
\newblock In \emph{International Conference on Dependable Systems and Networks
  (DSN'06)}, pages 467--472.

\bibitem[{Mohammad et~al.(2018)Mohammad, Bravo-Marquez, Salameh, and
  Kiritchenko}]{mohammad2018semeval}
Saif Mohammad, Felipe Bravo-Marquez, Mohammad Salameh, and Svetlana
  Kiritchenko. 2018.
\newblock {Semeval-2018 Task 1: Affect in Tweets}.
\newblock In \emph{Proceedings of the 12th International Workshop on Semantic
  Evaluation}, pages 1--17.
\newblock New Orleans, LA, USA.

\bibitem[{Mohammad et~al.(2016)Mohammad, Kiritchenko, Sobhani, Zhu, and
  Cherry}]{mohammad2016semeval}
Saif Mohammad, Svetlana Kiritchenko, Parinaz Sobhani, Xiaodan Zhu, and Colin
  Cherry. 2016.
\newblock {Semeval-2016 Task 6: Detecting Stance in Tweets}.
\newblock In \emph{Proceedings of the 10th International Workshop on Semantic
  Evaluation (SemEval-2016)}, pages 31--41.
\newblock San Diego, California, USA.

\bibitem[{Mollas et~al.(2022)Mollas, Chrysopoulou, Karlos, and
  Tsoumakas}]{mollasETHOSMultiLabelHate2022}
Ioannis Mollas, Zoe Chrysopoulou, Stamatis Karlos, and Grigorios Tsoumakas.
  2022.
\newblock \href {https://doi.org/10.1007/s40747-021-00608-2} {{{ETHOS}}: {{A
  Multi-Label Hate Speech Detection Dataset}}}.
\newblock \emph{Complex \& Intelligent Systems}.

\bibitem[{Nakov et~al.(2021)Nakov, Nayak, Dent, Bhatawdekar, Sarwar, Hardalov,
  Dinkov, Zlatkova, Bouchard, and Augenstein}]{nakov_detecting_2021}
Preslav Nakov, Vibha Nayak, Kyle Dent, Ameya Bhatawdekar, Sheikh~Muhammad
  Sarwar, Momchil Hardalov, Yoan Dinkov, Dimitrina Zlatkova, Guillaume
  Bouchard, and Isabelle Augenstein. 2021.
\newblock \href {http://arxiv.org/abs/2103.00153} {Detecting {Abusive}
  {Language} on {Online} {Platforms}: {A} {Critical} {Analysis}}.
\newblock \emph{arXiv:2103.00153 [cs]}.
\newblock ArXiv: 2103.00153.

\bibitem[{Parikh et~al.(2019)Parikh, Abburi, Badjatiya, Krishnan, Chhaya,
  Gupta, and Varma}]{parikh-etal-2019-multi}
Pulkit Parikh, Harika Abburi, Pinkesh Badjatiya, Radhika Krishnan, Niyati
  Chhaya, Manish Gupta, and Vasudeva Varma. 2019.
\newblock \href {https://doi.org/10.18653/v1/D19-1174} {Multi-label
  categorization of accounts of sexism using a neural framework}.
\newblock In \emph{Proceedings of the 2019 Conference on Empirical Methods in
  Natural Language Processing and the 9th International Joint Conference on
  Natural Language Processing (EMNLP-IJCNLP)}, pages 1642--1652, Hong Kong,
  China. Association for Computational Linguistics.

\bibitem[{Plaza-del Arco et~al.(2021)Plaza-del Arco, Molina-Gonz{\'a}lez,
  L{\'o}pez, and Mart{\'\i}n-Valdivia}]{plaza2021sexism}
Flor~Miriam Plaza-del Arco, M~Dolores Molina-Gonz{\'a}lez, LAU L{\'o}pez, and
  MT~Mart{\'\i}n-Valdivia. 2021.
\newblock Sexism identification in social networks using a multi-task learning
  system.
\newblock In \emph{Proceedings of the Iberian Languages Evaluation Forum
  (IberLEF 2021) co-located with the Conference of the Spanish Society for
  Natural Language Processing (SEPLN 2021), XXXVII International Conference of
  the Spanish Society for Natural Language Processing., M{\'a}laga, Spain},
  volume 2943, pages 491--499.

\bibitem[{Pradhan et~al.(2020)Pradhan, Chaturvedi, Tripathi, and
  Sharma}]{pradhan_review_2020}
Rahul Pradhan, Ankur Chaturvedi, Aprna Tripathi, and Dilip~Kumar Sharma. 2020.
\newblock \href {https://doi.org/10.1007/978-981-15-0694-9_41} {A {Review} on
  {Offensive} {Language} {Detection}}.
\newblock In \emph{Advances in {Data} and {Information} {Sciences}}, Lecture
  {Notes} in {Networks} and {Systems}, pages 433--439, Singapore. Springer.

\bibitem[{Raffel et~al.(2020)Raffel, Shazeer, Roberts, Lee, Narang, Matena,
  Zhou, Li, and Liu}]{raffel_exploring_2020}
Colin Raffel, Noam Shazeer, Adam Roberts, Katherine Lee, Sharan Narang, Michael
  Matena, Yanqi Zhou, Wei Li, and Peter~J. Liu. 2020.
\newblock \href {http://jmlr.org/papers/v21/20-074.html} {Exploring the
  {Limits} of {Transfer} {Learning} with a {Unified} {Text}-to-{Text}
  {Transformer}}.
\newblock \emph{Journal of Machine Learning Research}, 21(140):1--67.

\bibitem[{Rodr{\i}guez-S{\'a}nchez et~al.(2021)Rodr{\i}guez-S{\'a}nchez,
  Carrillo-de Albornoz, and Plaza}]{rodriguez2021multi}
Francisco Rodr{\i}guez-S{\'a}nchez, Jorge Carrillo-de Albornoz, and Laura
  Plaza. 2021.
\newblock A multi-task and multilingual model for sexism identification in
  social networks.
\newblock \emph{Proceedings of the Iberian Languages Evaluation Forum (IberLEF
  2021) co-located with the Conference of the Spanish Society for Natural
  Language Processing (SEPLN 2021)}.

\bibitem[{Rodr{\'i}guez‐S{\'a}nchez et~al.(2021)Rodr{\'i}guez‐S{\'a}nchez,
  de~Albornoz, Plaza, Gonzalo, Rosso, Comet, and
  Donoso}]{RodrguezSnchez2021OverviewOE}
Francisco Rodr{\'i}guez‐S{\'a}nchez, Jorge~Carrillo de~Albornoz, Laura Plaza,
  Julio Gonzalo, Paolo Rosso, Miriam Comet, and Trinidad Donoso. 2021.
\newblock Overview of exist 2021: sexism identification in social networks.
\newblock \emph{Proces. del Leng. Natural}, 69:229--240.

\bibitem[{Rosenthal et~al.(2017)Rosenthal, Farra, and
  Nakov}]{rosenthal-etal-2017-semeval}
Sara Rosenthal, Noura Farra, and Preslav Nakov. 2017.
\newblock \href {https://doi.org/10.18653/v1/S17-2088} {{S}em{E}val-2017 task
  4: Sentiment analysis in {T}witter}.
\newblock In \emph{Proceedings of the 11th International Workshop on Semantic
  Evaluation ({S}em{E}val-2017)}, pages 502--518, Vancouver, Canada.
  Association for Computational Linguistics.

\bibitem[{Ruder(2017)}]{ruder2017overview}
Sebastian Ruder. 2017.
\newblock An overview of multi-task learning in deep neural networks.
\newblock \emph{arXiv preprint arXiv:1706.05098}.

\bibitem[{Safi~Samghabadi et~al.(2020)Safi~Samghabadi, Patwa, PYKL, Mukherjee,
  Das, and Solorio}]{safi-samghabadi-etal-2020-aggression}
Niloofar Safi~Samghabadi, Parth Patwa, Srinivas PYKL, Prerana Mukherjee,
  Amitava Das, and Thamar Solorio. 2020.
\newblock \href {https://aclanthology.org/2020.trac-1.20} {Aggression and
  misogyny detection using {BERT}: A multi-task approach}.
\newblock In \emph{Proceedings of the Second Workshop on Trolling, Aggression
  and Cyberbullying}, pages 126--131, Marseille, France. European Language
  Resources Association (ELRA).

\bibitem[{Sap et~al.(2020)Sap, Gabriel, Qin, Jurafsky, Smith, and
  Choi}]{sap-etal-2020-social}
Maarten Sap, Saadia Gabriel, Lianhui Qin, Dan Jurafsky, Noah~A. Smith, and
  Yejin Choi. 2020.
\newblock \href {https://doi.org/10.18653/v1/2020.acl-main.486} {Social bias
  frames: Reasoning about social and power implications of language}.
\newblock In \emph{Proceedings of the 58th Annual Meeting of the Association
  for Computational Linguistics}, pages 5477--5490, Online. Association for
  Computational Linguistics.

\bibitem[{Simons(2015)}]{simons2015addressing}
Rachel~Noelle Simons. 2015.
\newblock A{ddressing Gender-Based Harassment in Social Media: A Call to
  Action}.
\newblock \emph{iConference 2015 Proceedings}.
\newblock Online.

\bibitem[{Sun et~al.(2021)Sun, Yang, Li, Zhang, Meng, Qiu, Wang, Hovy, and
  Li}]{sun_interpreting_2021}
Xiaofei Sun, Diyi Yang, Xiaoya Li, Tianwei Zhang, Yuxian Meng, Han Qiu, Guoyin
  Wang, Eduard Hovy, and Jiwei Li. 2021.
\newblock \href {http://arxiv.org/abs/2110.10470} {{Interpreting Deep Learning
  Models in Natural Language Processing: A Review}}.
\newblock Arxiv:2110.10470.

\bibitem[{Van~Hee et~al.(2018)Van~Hee, Lefever, and
  Hoste}]{van-hee-etal-2018-semeval}
Cynthia Van~Hee, Els Lefever, and V{\'e}ronique Hoste. 2018.
\newblock \href {https://doi.org/10.18653/v1/S18-1005} {{S}em{E}val-2018 task
  3: Irony detection in {E}nglish tweets}.
\newblock In \emph{Proceedings of the 12th International Workshop on Semantic
  Evaluation}, pages 39--50, New Orleans, Louisiana. Association for
  Computational Linguistics.

\bibitem[{Vidgen et~al.(2021)Vidgen, Thrush, Waseem, and
  Kiela}]{vidgen-etal-2021-learning}
Bertie Vidgen, Tristan Thrush, Zeerak Waseem, and Douwe Kiela. 2021.
\newblock \href {https://doi.org/10.18653/v1/2021.acl-long.132} {Learning from
  the worst: Dynamically generated datasets to improve online hate detection}.
\newblock In \emph{Proceedings of the 59th Annual Meeting of the Association
  for Computational Linguistics and the 11th International Joint Conference on
  Natural Language Processing (Volume 1: Long Papers)}, pages 1667--1682,
  Online. Association for Computational Linguistics.

\bibitem[{Wang et~al.(2021)Wang, Fang, Khabsa, Mao, and
  Ma}]{wang_entailment_2021}
Sinong Wang, Han Fang, Madian Khabsa, Hanzi Mao, and Hao Ma. 2021.
\newblock \href {http://arxiv.org/abs/2104.14690} {Entailment as {Few}-{Shot}
  {Learner}}.
\newblock \emph{arXiv:2104.14690 [cs]}.
\newblock ArXiv: 2104.14690.

\bibitem[{Waseem(2016)}]{waseem-2016-racist}
Zeerak Waseem. 2016.
\newblock \href {https://doi.org/10.18653/v1/W16-5618} {Are you a racist or am
  {I} seeing things? annotator influence on hate speech detection on
  {T}witter}.
\newblock In \emph{Proceedings of the First Workshop on {NLP} and Computational
  Social Science}, pages 138--142, Austin, Texas. Association for Computational
  Linguistics.

\bibitem[{Waseem and Hovy(2016)}]{waseem_hateful_2016}
Zeerak Waseem and Dirk Hovy. 2016.
\newblock \href {https://doi.org/10.18653/v1/N16-2013} {Hateful {Symbols} or
  {Hateful} {People}? {Predictive} {Features} for {Hate} {Speech} {Detection}
  on {Twitter}}.
\newblock In \emph{Proceedings of the {NAACL} {Student} {Research} {Workshop}},
  pages 88--93, San Diego, California. Association for Computational
  Linguistics.

\bibitem[{Zampieri et~al.(2019)Zampieri, Malmasi, Nakov, Rosenthal, Farra, and
  Kumar}]{zampieri2019semeval}
Marcos Zampieri, Shervin Malmasi, Preslav Nakov, Sara Rosenthal, Noura Farra,
  and Ritesh Kumar. 2019.
\newblock Semeval-2019 task 6: Identifying and categorizing offensive language
  in social media (offenseval).
\newblock In \emph{Proceedings of the 13th International Workshop on Semantic
  Evaluation}, pages 75--86.

\bibitem[{Zhang et~al.(2022)Zhang, Wang, Yang, Yu, Vu, and
  Lei}]{Zhang2022LongtailedEM}
Ruohong Zhang, Yau-Shian Wang, Yiming Yang, Donghan Yu, Tom Vu, and Li~Lei.
  2022.
\newblock Long-tailed extreme multi-label text classification with generated
  pseudo label descriptions.
\newblock \emph{ArXiv}, abs/2204.00958.

\end{thebibliography}
\bibliographystyle{acl_natbib}

\appendix

\newpage
\begin{figure*}
\section{Full Results}
\label{sec:app:full-results}
\nopagebreak[1]
\centering
\setlength{\tabcolsep}{9pt}
\renewcommand\figurename{Table}
\renewcommand{\thefigure}{6}
\resizebox{0.95\textwidth}{!}{
\begin{tabular}{lrccl:cccr:ccr}
\textbf{} & \multicolumn{1}{l}{\textbf{}} & \multicolumn{1}{l}{} & \multicolumn{1}{l}{} &  & \multicolumn{4}{c:}{Development set} & \multicolumn{3}{c}{Test set} \\
Training sets and training phases & \multicolumn{1}{l}{Run} & LD & DI & Base model & A & $\rho$ & B & \multicolumn{1}{c:}{C} & A & B & \multicolumn{1}{c}{C} \\
\toprule
EDOS A & 1 & \xmark & \xmark & \textbf{DBV3} & \multicolumn{1}{r}{0.840} & \multicolumn{1}{r}{0.7} & - & \multicolumn{1}{c:}{-} & \multicolumn{1}{r}{0.837} & - & \multicolumn{1}{c}{-} \\
EDOS A & 2 & \xmark & \xmark & \textbf{DBV3} & \multicolumn{1}{r}{0.845} & \multicolumn{1}{r}{0.5} & - & \multicolumn{1}{c:}{-} & \multicolumn{1}{r}{0.848} & - & \multicolumn{1}{c}{-} \\
EDOS A & 3 & \xmark & \xmark & \textbf{DBV3} & \multicolumn{1}{r}{0.837} & \multicolumn{1}{r}{0.5} & - & \multicolumn{1}{c:}{-} & \multicolumn{1}{r}{0.836} & - & \multicolumn{1}{c}{-} \\
EDOS B & 1 & \xmark & \xmark & \textbf{DBV3} & - & - & 0.159 & \multicolumn{1}{c:}{-} & - & 0.159 & \multicolumn{1}{c}{-} \\
EDOS B & 2 & \xmark & \xmark & \textbf{DBV3} & - & - & 0.159 & \multicolumn{1}{c:}{-} & - & 0.159 & \multicolumn{1}{c}{-} \\
EDOS B & 3 & \xmark & \xmark & \textbf{DBV3} & - & - & 0.306 & \multicolumn{1}{c:}{-} & - & 0.287 & \multicolumn{1}{c}{-} \\
EDOS C & 1 & \xmark & \xmark & \textbf{DBV3} & - & - & - & 0.114 & - & - & 0.115 \\
EDOS C & 2 & \xmark & \xmark & \textbf{DBV3} & - & - & - & 0.136 & - & - & 0.126 \\
EDOS C & 3 & \xmark & \xmark & \textbf{DBV3} & - & - & - & 0.110 & - & - & 0.124 \\
\hdashline
EDOS A & 1 & \cmark & \xmark & \textbf{DBV3} & \multicolumn{1}{r}{0.853} & \multicolumn{1}{r}{0.5} & - & \multicolumn{1}{c:}{-} & \multicolumn{1}{r}{0.852} & - & \multicolumn{1}{c}{-} \\
EDOS A & 2 & \cmark & \xmark & \textbf{DBV3} & \multicolumn{1}{r}{0.845} & \multicolumn{1}{r}{0.5} & - & \multicolumn{1}{c:}{-} & \multicolumn{1}{r}{0.852} & - & \multicolumn{1}{c}{-} \\
EDOS A & 3 & \cmark & \xmark & \textbf{DBV3} & \multicolumn{1}{r}{0.851} & \multicolumn{1}{r}{0.5} & - & \multicolumn{1}{c:}{-} & \multicolumn{1}{r}{0.849} & - & \multicolumn{1}{c}{-} \\
EDOS B & 1 & \cmark & \xmark & \textbf{DBV3} & - & - & \multicolumn{1}{r}{0.162} & \multicolumn{1}{c:}{-} & - & \multicolumn{1}{r}{0.159} & \multicolumn{1}{c}{-} \\
EDOS B & 2 & \cmark & \xmark & \textbf{DBV3} & - & - & \multicolumn{1}{r}{0.162} & \multicolumn{1}{c:}{-} & - & \multicolumn{1}{r}{0.159} & \multicolumn{1}{c}{-} \\
EDOS B & 3 & \cmark & \xmark & \textbf{DBV3} & - & - & \multicolumn{1}{r}{0.159} & \multicolumn{1}{c:}{-} & - & \multicolumn{1}{r}{0.161} & \multicolumn{1}{c}{-} \\
EDOS C & 1 & \cmark & \xmark & \textbf{DBV3} & - & - & - & 0.117 & - & - & 0.118 \\
EDOS C & 2 & \cmark & \xmark & \textbf{DBV3} & - & - & - & 0.094 & - & - & 0.086 \\
EDOS C & 3 & \cmark & \xmark & \textbf{DBV3} & - & - & - & 0.078 & - & - & 0.089 \\
\hdashline
EDOS ABC & 1 & \cmark & \xmark & \textbf{DBV3} & \multicolumn{1}{r}{0.865} & \multicolumn{1}{r}{0.5} & \multicolumn{1}{r}{0.556} & 0.225 & \multicolumn{1}{r}{0.851} & \multicolumn{1}{r}{0.530} & 0.253 \\
EDOS ABC & 2 & \cmark & \xmark & \textbf{DBV3} & \multicolumn{1}{r}{0.850} & \multicolumn{1}{r}{0.5} & \multicolumn{1}{r}{0.466} & 0.193 & \multicolumn{1}{r}{0.845} & \multicolumn{1}{r}{0.449} & 0.184 \\
EDOS ABC & 3 & \cmark & \xmark & \textbf{DBV3} & \multicolumn{1}{r}{0.860} & \multicolumn{1}{r}{0.7} & \multicolumn{1}{r}{0.570} & 0.329 & \multicolumn{1}{r}{0.857} & \multicolumn{1}{r}{0.533} & 0.309 \\
EDOS ABC & 1 & \cmark & \xmark & \textbf{\textbf{DBV3}-NLI} & \multicolumn{1}{r}{0.855} & \multicolumn{1}{r}{0.5} & \multicolumn{1}{r}{0.616} & 0.439 & \multicolumn{1}{r}{0.854} & \multicolumn{1}{r}{0.554} & 0.350 \\
EDOS ABC & 2 & \cmark & \xmark & \textbf{\textbf{DBV3}-NLI} & \multicolumn{1}{r}{0.855} & \multicolumn{1}{r}{0.5} & \multicolumn{1}{r}{0.617} & 0.431 & \multicolumn{1}{r}{0.852} & \multicolumn{1}{r}{0.555} & 0.355 \\
EDOS ABC & 3 & \cmark & \xmark & \textbf{\textbf{DBV3}-NLI} & \multicolumn{1}{r}{0.854} & \multicolumn{1}{r}{0.6} & \multicolumn{1}{r}{0.614} & 0.440 & \multicolumn{1}{r}{0.855} & \multicolumn{1}{r}{0.558} & 0.351 \\
\hdashline
Ph1: EDOS ABC, Ph2: EDOS A & 1 & \cmark & \xmark & \textbf{\textbf{DBV3}-NLI} & \multicolumn{1}{r}{0.853} & \multicolumn{1}{r}{0.5} & - & \multicolumn{1}{c:}{-} & \multicolumn{1}{r}{0.848} & - & \multicolumn{1}{c}{-} \\
Ph1: EDOS ABC, Ph2: EDOS A & 2 & \cmark & \xmark & \textbf{\textbf{DBV3}-NLI} & \multicolumn{1}{r}{0.855} & \multicolumn{1}{r}{0.6} & - & \multicolumn{1}{c:}{-} & \multicolumn{1}{r}{0.848} & - & \multicolumn{1}{c}{-} \\
Ph1: EDOS ABC, Ph2: EDOS A & 3 & \cmark & \xmark & \textbf{\textbf{DBV3}-NLI} & \multicolumn{1}{r}{0.854} & \multicolumn{1}{r}{0.9} & - & \multicolumn{1}{c:}{-} & \multicolumn{1}{r}{0.855} & - & \multicolumn{1}{c}{-} \\
Ph1: EDOS ABC, Ph2: EDOS B & 1 & \cmark & \xmark & \textbf{\textbf{DBV3}-NLI} & - & - & \multicolumn{1}{r}{0.665} & \multicolumn{1}{c:}{-} & - & \multicolumn{1}{r}{0.615} & \multicolumn{1}{c}{-} \\
Ph1: EDOS ABC, Ph2: EDOS B & 2 & \cmark & \xmark & \textbf{\textbf{DBV3}-NLI} & - & - & \multicolumn{1}{r}{0.683} & \multicolumn{1}{c:}{-} & - & \multicolumn{1}{r}{0.637} & \multicolumn{1}{c}{-} \\
Ph1: EDOS ABC, Ph2: EDOS B & 3 & \cmark & \xmark & \textbf{\textbf{DBV3}-NLI} & - & - & \multicolumn{1}{r}{0.683} & \multicolumn{1}{c:}{-} & - & \multicolumn{1}{r}{0.616} & \multicolumn{1}{c}{-} \\
Ph1: EDOS ABC, Ph2: EDOS C & 1 & \cmark & \xmark & \textbf{\textbf{DBV3}-NLI} & - & - & - & 0.496 & - & - & 0.412 \\
Ph1: EDOS ABC, Ph2: EDOS C & 2 & \cmark & \xmark & \textbf{\textbf{DBV3}-NLI} & - & - & - & 0.496 & - & - & 0.412 \\
Ph1: EDOS ABC, Ph2: EDOS C & 3 & \cmark & \xmark & \textbf{\textbf{DBV3}-NLI} & - & - & - & 0.496 & - & - & 0.412 \\
\hdashline
Ph1: AUX + EDOS ABC & 1 & \cmark & \xmark & \textbf{\textbf{DBV3}-NLI} & \multicolumn{1}{r}{0.825} & \multicolumn{1}{r}{0.5} & \multicolumn{1}{r}{0.283} & 0.247 & \multicolumn{1}{r}{0.831} & \multicolumn{1}{r}{0.263} & 0.228 \\
Ph1: AUX + EDOS ABC & 2 & \cmark & \xmark & \textbf{\textbf{DBV3}-NLI} & \multicolumn{1}{r}{0.825} & \multicolumn{1}{r}{0.5} & \multicolumn{1}{r}{0.291} & 0.230 & \multicolumn{1}{r}{0.828} & \multicolumn{1}{r}{0.269} & 0.237 \\
Ph1: AUX + EDOS ABC & 3 & \cmark & \xmark & \textbf{\textbf{DBV3}-NLI} & \multicolumn{1}{r}{0.824} & \multicolumn{1}{r}{0.5} & \multicolumn{1}{r}{0.302} & 0.245 & \multicolumn{1}{r}{0.827} & \multicolumn{1}{r}{0.284} & 0.239 \\
\hdashline
Ph1: AUX + EDOS ABC Ph2: EDOS ABC & 1 & \cmark & \xmark & \textbf{\textbf{DBV3}-NLI} & \multicolumn{1}{r}{0.850} & \multicolumn{1}{r}{0.6} & \multicolumn{1}{r}{0.601} & 0.422 & \multicolumn{1}{r}{0.860} & \multicolumn{1}{r}{0.541} & 0.382 \\
Ph1: AUX + EDOS ABC Ph2: EDOS ABC & 2 & \cmark & \xmark & \textbf{\textbf{DBV3}-NLI} & \multicolumn{1}{r}{0.851} & \multicolumn{1}{r}{0.6} & \multicolumn{1}{r}{0.608} & 0.421 & \multicolumn{1}{r}{0.859} & \multicolumn{1}{r}{0.549} & 0.379 \\
Ph1: AUX + EDOS ABC Ph2: EDOS ABC & 3 & \cmark & \xmark & \textbf{\textbf{DBV3}-NLI} & \multicolumn{1}{r}{0.853} & \multicolumn{1}{r}{0.5} & \multicolumn{1}{r}{0.602} & 0.424 & \multicolumn{1}{r}{0.857} & \multicolumn{1}{r}{0.538} & 0.380 \\
\hdashline
Ph1: AUX + EDOS ABC Ph2: EDOS ABC, Ph3: EDOS A & 1 & \cmark & \xmark & \textbf{\textbf{DBV3}-NLI} & \multicolumn{1}{r}{0.855} & \multicolumn{1}{r}{0.6} & - & \multicolumn{1}{c:}{-} & \multicolumn{1}{r}{0.860} & - & \multicolumn{1}{c}{-} \\
Ph1: AUX + EDOS ABC Ph2: EDOS ABC, Ph3: EDOS A & 2 & \cmark & \xmark & \textbf{\textbf{DBV3}-NLI} & \multicolumn{1}{r}{0.854} & \multicolumn{1}{r}{0.7} & - & \multicolumn{1}{c:}{-} & \multicolumn{1}{r}{0.858} & - & \multicolumn{1}{c}{-} \\
Ph1: AUX + EDOS ABC Ph2: EDOS ABC, Ph3: EDOS A & 3 & \cmark & \xmark & \textbf{\textbf{DBV3}-NLI} & \multicolumn{1}{r}{0.855} & \multicolumn{1}{r}{0.7} & - & \multicolumn{1}{c:}{-} & \multicolumn{1}{r}{0.857} & - & \multicolumn{1}{c}{-} \\
Ph1: AUX + EDOS ABC Ph2: EDOS ABC, Ph3: EDOS B & 1 & \cmark & \xmark & \textbf{\textbf{DBV3}-NLI} & - & - & \multicolumn{1}{r}{0.663} & \multicolumn{1}{c:}{-} & - & \multicolumn{1}{r}{0.594} & \multicolumn{1}{c}{-} \\
Ph1: AUX + EDOS ABC Ph2: EDOS ABC, Ph3: EDOS B & 2 & \cmark & \xmark & \textbf{\textbf{DBV3}-NLI} & - & - & \multicolumn{1}{r}{0.693} & \multicolumn{1}{c:}{-} & - & \multicolumn{1}{r}{0.657} & \multicolumn{1}{c}{-} \\
Ph1: AUX + EDOS ABC Ph2: EDOS ABC, Ph3: EDOS B & 3 & \cmark & \xmark & \textbf{\textbf{DBV3}-NLI} & - & - & \multicolumn{1}{r}{0.689} & \multicolumn{1}{c:}{-} & - & \multicolumn{1}{r}{0.649} & \multicolumn{1}{c}{-} \\
Ph1: AUX + EDOS ABC Ph2: EDOS ABC, Ph3: EDOS C & 1 & \cmark & \xmark & \textbf{\textbf{DBV3}-NLI} & - & - & - & 0.507 & - & - & 0.423 \\
Ph1: AUX + EDOS ABC Ph2: EDOS ABC, Ph3: EDOS C & 2 & \cmark & \xmark & \textbf{\textbf{DBV3}-NLI} & - & - & - & 0.495 & - & - & 0.425 \\
Ph1: AUX + EDOS ABC Ph2: EDOS ABC, Ph3: EDOS C & 3 & \cmark & \xmark & \textbf{\textbf{DBV3}-NLI} & - & - & - & 0.478 & - & - & 0.401 \\
\hdashline
Ph1: AUX + EDOS ABC & 1 & \cmark & \cmark & \textbf{\textbf{DBV3}-NLI} & \multicolumn{1}{r}{0.796} & \multicolumn{1}{r}{0.5} & \multicolumn{1}{r}{0.260} & 0.196 & \multicolumn{1}{r}{0.805} & \multicolumn{1}{r}{0.236} & 0.214 \\
Ph1: AUX + EDOS ABC & 2 & \cmark & \cmark & \textbf{\textbf{DBV3}-NLI} & \multicolumn{1}{r}{0.798} & \multicolumn{1}{r}{0.5} & \multicolumn{1}{r}{0.259} & 0.206 & \multicolumn{1}{r}{0.801} & \multicolumn{1}{r}{0.237} & 0.207 \\
Ph1: AUX + EDOS ABC & 3 & \cmark & \cmark & \textbf{\textbf{DBV3}-NLI} & \multicolumn{1}{r}{0.793} & \multicolumn{1}{r}{0.5} & \multicolumn{1}{r}{0.257} & 0.226 & \multicolumn{1}{r}{0.803} & \multicolumn{1}{r}{0.253} & 0.230 \\
\hdashline
Ph1: AUX + EDOS ABC Ph2: EDOS ABC & 1 & \cmark & \cmark & \textbf{\textbf{DBV3}-NLI} & \multicolumn{1}{r}{0.862} & \multicolumn{1}{r}{0.6} & \multicolumn{1}{r}{0.637} & 0.466 & \multicolumn{1}{r}{0.859} & \multicolumn{1}{r}{0.605} & 0.395 \\
Ph1: AUX + EDOS ABC Ph2: EDOS ABC & 2 & \cmark & \cmark & \textbf{\textbf{DBV3}-NLI} & \multicolumn{1}{r}{0.852} & \multicolumn{1}{r}{0.5} & \multicolumn{1}{r}{0.565} & 0.411 & \multicolumn{1}{r}{0.857} & \multicolumn{1}{r}{0.533} & 0.370 \\
Ph1: AUX + EDOS ABC Ph2: EDOS ABC & 3 & \cmark & \cmark & \textbf{\textbf{DBV3}-NLI} & \multicolumn{1}{r}{0.849} & \multicolumn{1}{r}{0.6} & \multicolumn{1}{r}{0.569} & 0.424 & \multicolumn{1}{r}{0.859} & \multicolumn{1}{r}{0.534} & 0.366 \\
\hdashline
\rowcolor[HTML]{D9D9D9} 
Ph1: AUX + EDOS ABC Ph2: EDOS ABC, Ph3: EDOS A & 1 & \cmark & \cmark & \textbf{\textbf{DBV3}-NLI} & \multicolumn{1}{r}{\cellcolor[HTML]{D9D9D9}0.858} & \multicolumn{1}{r}{\cellcolor[HTML]{D9D9D9}0.7} & - & \multicolumn{1}{c:}{\cellcolor[HTML]{D9D9D9}-} & \multicolumn{1}{r}{\cellcolor[HTML]{D9D9D9}0.859} & - & \multicolumn{1}{c}{\cellcolor[HTML]{D9D9D9}-} \\
\rowcolor[HTML]{D9D9D9} 
Ph1: AUX + EDOS ABC Ph2: EDOS ABC, Ph3: EDOS A & 2 & \cmark & \cmark & \textbf{\textbf{DBV3}-NLI} & \multicolumn{1}{r}{\cellcolor[HTML]{D9D9D9}0.855} & \multicolumn{1}{r}{\cellcolor[HTML]{D9D9D9}0.6} & - & \multicolumn{1}{c:}{\cellcolor[HTML]{D9D9D9}-} & \multicolumn{1}{r}{\cellcolor[HTML]{D9D9D9}0.856} & - & \multicolumn{1}{c}{\cellcolor[HTML]{D9D9D9}-} \\
\rowcolor[HTML]{D9D9D9} 
Ph1: AUX + EDOS ABC Ph2: EDOS ABC, Ph3: EDOS A & 3 & \cmark & \cmark & \textbf{\textbf{DBV3}-NLI} & \multicolumn{1}{r}{\cellcolor[HTML]{D9D9D9}0.862} & \multicolumn{1}{r}{\cellcolor[HTML]{D9D9D9}0.5} & - & \multicolumn{1}{c:}{\cellcolor[HTML]{D9D9D9}-} & \multicolumn{1}{r}{\cellcolor[HTML]{D9D9D9}0.861} & - & \multicolumn{1}{c}{\cellcolor[HTML]{D9D9D9}-} \\
Ph1: AUX + EDOS ABC Ph2: EDOS ABC, Ph3: EDOS B & 1 & \cmark & \cmark & \textbf{\textbf{DBV3}-NLI} & - & - & \multicolumn{1}{r}{0.674} & \multicolumn{1}{c:}{-} & - & \multicolumn{1}{r}{0.633} & \multicolumn{1}{c}{-} \\
Ph1: AUX + EDOS ABC Ph2: EDOS ABC, Ph3: EDOS B & 2 & \cmark & \cmark & \textbf{\textbf{DBV3}-NLI} & - & - & \multicolumn{1}{r}{0.665} & \multicolumn{1}{c:}{-} & - & \multicolumn{1}{r}{0.642} & \multicolumn{1}{c}{-} \\
Ph1: AUX + EDOS ABC Ph2: EDOS ABC, Ph3: EDOS B & 3 & \cmark & \cmark & \textbf{\textbf{DBV3}-NLI} & - & - & \multicolumn{1}{r}{0.664} & \multicolumn{1}{c:}{-} & - & \multicolumn{1}{r}{0.613} & \multicolumn{1}{c}{-} \\
Ph1: AUX + EDOS ABC Ph2: EDOS ABC, Ph3: EDOS C & 1 & \cmark & \cmark & \textbf{\textbf{DBV3}-NLI} & - & - & - & 0.522 & - & - & 0.455 \\
Ph1: AUX + EDOS ABC Ph2: EDOS ABC, Ph3: EDOS C & 2 & \cmark & \cmark & \textbf{\textbf{DBV3}-NLI} & - & - & - & 0.464 & - & - & 0.419 \\
Ph1: AUX + EDOS ABC Ph2: EDOS ABC, Ph3: EDOS C & 3 & \cmark & \cmark & \textbf{\textbf{DBV3}-NLI} & - & - & - & 0.473 & - & - & 0.419 \\
\rowcolor[HTML]{FFFFFF} 
\hdashline
\rowcolor[HTML]{D9D9D9} 
Ph1: AUX + EDOS ABC Ph2: EDOS ABC, Ph3: EDOS B (up to \textasciitilde19\%) & 1 & \cmark & \cmark & \textbf{\textbf{DBV3}-NLI} & - & - & 0.679 & \multicolumn{1}{c:}{\cellcolor[HTML]{D9D9D9}-} & - & 0.653 & \multicolumn{1}{c}{\cellcolor[HTML]{D9D9D9}-} \\
\rowcolor[HTML]{D9D9D9} 
Ph1: AUX + EDOS ABC Ph2: EDOS ABC, Ph3: EDOS B (up to \textasciitilde19\%) & 2 & \cmark & \cmark & \textbf{\textbf{DBV3}-NLI} & - & - & 0.677 & \multicolumn{1}{c:}{\cellcolor[HTML]{D9D9D9}-} & - & 0.642 & \multicolumn{1}{c}{\cellcolor[HTML]{D9D9D9}-} \\
\rowcolor[HTML]{D9D9D9} 
Ph1: AUX + EDOS ABC Ph2: EDOS ABC, Ph3: EDOS B (up to \textasciitilde19\%) & 3 & \cmark & \cmark & \textbf{\textbf{DBV3}-NLI} & - & - & 0.661 & \multicolumn{1}{c:}{\cellcolor[HTML]{D9D9D9}-} & - & 0.632 & \multicolumn{1}{c}{\cellcolor[HTML]{D9D9D9}-} \\
\rowcolor[HTML]{FFFFFF} 
\hdashline
\rowcolor[HTML]{D9D9D9} 
Ph1: AUX + EDOS ABC Ph2: EDOS ABC, Ph3: EDOS C (up to \textasciitilde9\%) & 1 & \cmark & \cmark & \textbf{\textbf{DBV3}-NLI} & - & - & - & 0.473 & - & - & 0.462 \\
\rowcolor[HTML]{D9D9D9} 
Ph1: AUX + EDOS ABC Ph2: EDOS ABC, Ph3: EDOS C (up to \textasciitilde9\%) & 2 & \cmark & \cmark & \textbf{\textbf{DBV3}-NLI} & - & - & - & 0.497 & - & - & 0.470 \\
\rowcolor[HTML]{D9D9D9} 
Ph1: AUX + EDOS ABC Ph2: EDOS ABC, Ph3: EDOS C (up to \textasciitilde9\%) & 3 & \cmark & \cmark & \textbf{\textbf{DBV3}-NLI} & - & - & - & 0.516 & - & - & 0.466 \\
\bottomrule
\end{tabular}
}
\label{tab:full-results}
\caption{Full results of the ablation study on the development and test set. \textit{LD} refers to label descriptions, and \textit{DI} refers to dataset identifiers. \textit{DBV3} refers to DeBERTa-V3-large and \textit{DBV3-NLI} refers to DeBERTa-V3-large fine-tuned on NLI datasets. $\rho$ refers to the threshold applied for subtask A. The settings containing the settings of the models submitted to the official evaluation are marked in grey. \textit{Ph1}, \textit{Ph2}, and \textit{Ph3} stand for training phase 1, 2, and 3 respectively.}
\end{figure*}
\end{document}